\documentclass[fleqn,10pt]{wlscirep}
\usepackage[utf8]{inputenc}
\usepackage[T1]{fontenc}
\usepackage{tabularx}
\usepackage{amsmath}
\title{A Comparative Study of Open-Source Large Language Models, GPT-4 and Claude 2: Multiple-Choice Test Taking
in Nephrology
}

\author[1]{Sean Wu}
\author[1]{Michael Koo}
\author[2]{Lesley Blum}
\author[2]{Andy Black}
\author[2]{Liyo Kao}
\author[1]{Fabien Scalzo}
\author[2,3,*]{Ira Kurtz}

\affil[1]{Keck Data Science Institute, Natural Science Department, Pepperdine University, Malibu, California, USA}
\affil[2]{Division of Nephrology, David Geffen School of Medicine, University of California, Los Angeles, California, USA}
\affil[3]{Brain Research Institute, University of California, Los Angeles, California, USA}

\affil[*]{ikurtz@mednet.ucla.edu}

\begin{abstract}
\textbf{Background:} In recent years, there have been significant breakthroughs in the field of natural language processing, particularly with the development of large language models (LLMs). These LLMs have showcased remarkable capabilities on various benchmarks. In the healthcare field, the exact role LLMs and other future AI models will play remains unclear. There is a potential for these models in the future to be used as part of adaptive physician training, medical co-pilot applications, and digital patient interaction scenarios. The ability of AI models to participate in medical training and patient care will depend in part on their mastery of the knowledge content of specific medical fields.  \\

\textbf{Methodology:} This study investigated the medical knowledge capability of LLMs, specifically in the context of internal medicine subspecialty multiple-choice test-taking ability. We compared the performance of several open-source LLMs (Koala 7B, Falcon 7B, Stable-Vicuna 13B, and Orca Mini 13B), to GPT-4 and Claude 2 on multiple-choice questions in the field of Nephrology. Nephrology was chosen as an example of a particularly conceptually complex subspecialty field within internal medicine. The study was conducted to evaluate the ability of LLM models to provide correct answers to nephSAP (Nephrology Self-Assessment Program) multiple-choice questions.\\

\textbf{Results:} The overall success of open-sourced LLMs in answering the 858 nephSAP multiple-choice questions correctly was 17.1\% – 25.5\%. In contrast, Claude 2 answered 54.4\% of the questions correctly, whereas GPT-4 achieved a score of 73.3\%. \\

\textbf{Conclusions:} We show that current widely used open-sourced LLMs do poorly in their ability for zero-shot reasoning when compared to GPT-4 and Claude 2. The findings of this study potentially have significant implications for the future of subspecialty medical training and patient care. Specifically, the demonstrated current superior capability of GPT-4 in accurately answering multiple-choice questions in Nephrology points to the utility of similar and more capable AI models in future medical applications that include multiple-choice question and answer test creation, adaptive individualized trainee teaching, medical co-pilot applications, and digital simulated physician-patient interactions in the outpatient and inpatient settings.

\end{abstract}

\begin{document}

\flushbottom
\maketitle
%
%
\thispagestyle{empty}

\section*{Introduction}

Large language models (LLMs) are artificial intelligence systems that use deep learning algorithms to understand and generate human-like natural language responses to a wide variety of prompts\cite{liu2023summary}. ChatGPT from OpenAI, which has
recently received much notoriety, is an LLM trained on a corpus of natural language text data coupled with supervised
learning, that has successfully taken advantage of the transformer algorithm breakthrough\cite{vaswani2017attention}. LLMs have turned out to be highly successful because of the emergent property that the texts they produce can be viewed in some sense as a “projection” of a human-like model of the external world. In the two months after its release in November 2022, approximately 100 million users signed up for ChatGPT. Its rapid acceptance is even more surprising, given that the technology is still in its infancy. 

One of the methods for assessing the capabilities of LLMs in knowledge-based fields is in their multiple choice test-taking ability\cite{nori2023capabilities}. In 2023, the release of GPT-4 by OpenAI took the world by storm with its impressive test-taking capabilities\cite{openai2023gpt4,nori2023capabilities}. Claude 2 from Anthropic, released in June 2023, has received increasing attention and has the ability to interpret larger input spaces (100k tokens). Recently, several open-source LLMs have been reported to be successful in different domains \cite{mukherjee2023orca,alpaca,seita-geng-koala,chiang2023vicuna}.

In the present study, we analyzed the overall ability of several open-source LLMs to answer Nephrology multiple-choice test questions successfully in comparison to GPT-4\cite{openai2023gpt4} and Claude 2\cite{claude-model-card}. We further assessed the capabilities of these models by determining their correct answer percentage in each of the various medical topics within Nephrology. Finally, we evaluated the open-sourced models by comparing the justification they gave for the answer they considered correct using the proximity of word embeddings in the vector space approach\cite{mikolov2013efficient}.

\section*{Methodology}
\subsection*{Large Language Models}
In this study, we evaluated the ability of several robust open-source LLM models including Koala\cite{seita-geng-koala}, Falcon\cite{penedo2023refinedweb}, Orca-mini\cite{orca-mini-13b} and Stable Vicuna\cite{chiang2023vicuna}, in addition to GPT-4 and Claude 2, to correctly answer multiple-choice questions in the field of Nephrology. Koala is a LLM that has undergone fine-tuning using the LLaMA network and is trained on ShareGPT and more high-quality datasets. Vicuna is another innovative model and was one of the first large-scale open-sourced models. While several LLMs models are trained on The Pile, an open-source language modeling data set, Falcon is trained on RefinedWeb, a corpus specifically curated to enhance scalability in language models. Currently, Falcon is the state-of-the-art open-source model, scoring 54.2 MMLU—an examination that tests the model's ability to solve problems in computer science, law, and other fields. Orca-mini (13B) is a replica LLM derived from the original Orca LLM. It is built using a model fine-tuned on the WizardLM, Alpaca, and Dolly datasets as per guidelines \cite{mukherjee2023orca}. Orca learns from a teacher LLM, gaining knowledge through explanations instead of just supervised pairs.

\begin{figure}[t]
\centering
\includegraphics[width=1\linewidth]{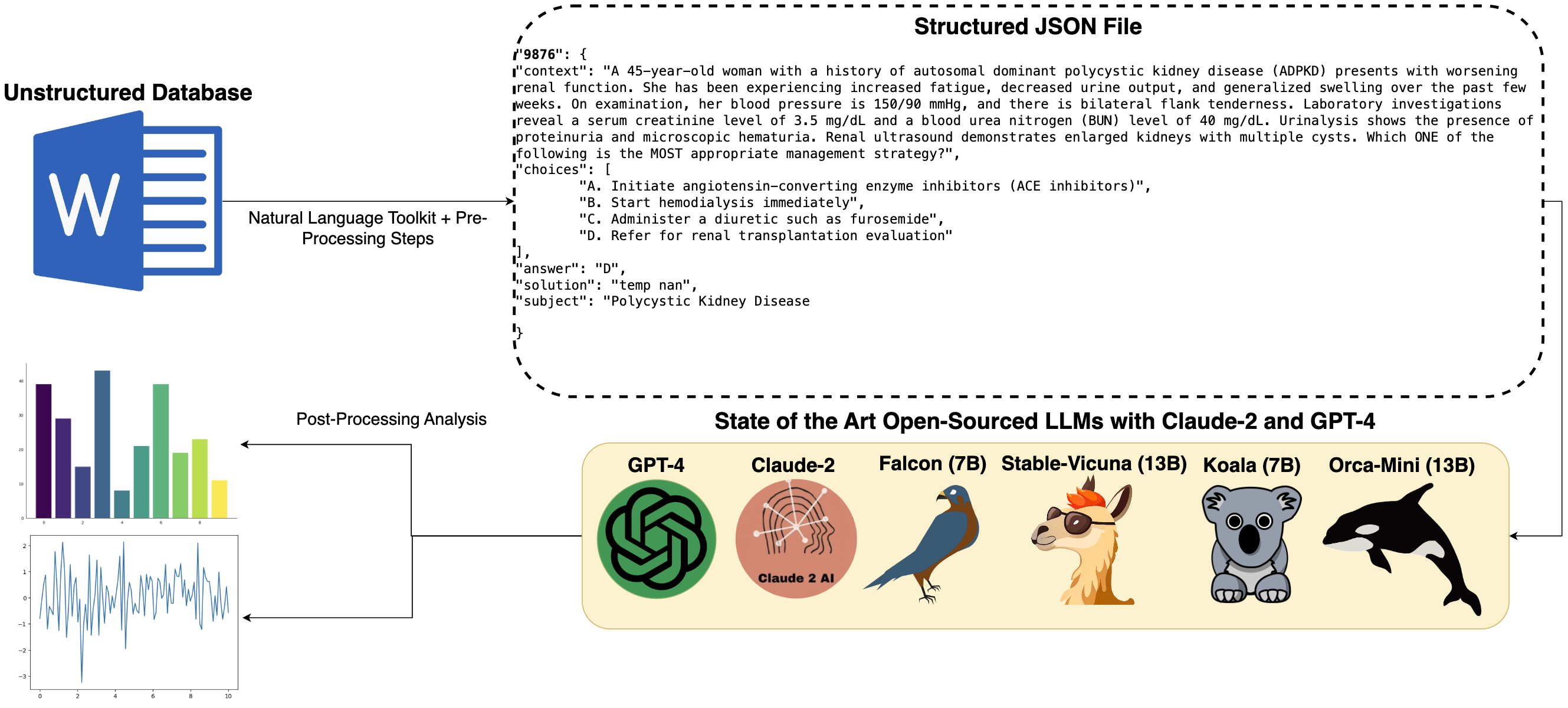}
\caption{Depiction of the workflow, encompassing data acquisition, pre-processing, and the utilization of LLMs for comparison purposes. The depicted structured json file was LLM generated as an example, and not part of the nephSAP question set.
}
\label{fig:stream}
\end{figure}

\subsection*{Dataset}
The dataset used as part of our experiments comprised 858 nephSAP (Nephrology Self-Assessment Program) multiple-choice questions and answers from January 2016 to April 2023. The format of the questions includes relevant background clinical information followed by a prompt for selecting the correct answer among the various choices. Questions with complex tables were omitted because of the difficulty encountered by the LLMs in their interpretation. The initial challenge entailed extracting a structured json file from an unstructured text file containing raw questions and answers. To accomplish this, we employed an automated parsing technique to extract the ``Questions" and ``Answers" from the file. Subsequently, we utilized the natural language processing toolkit (NLTK)\cite{loper2002nltk} to tokenize the text and accurately generate the json file (Figure 1). Finally, we traversed through a separate csv file that contained our ground truth answers in order to incorporate the correct answers acquired from the test bank. As a result, each example in our structured json file encompassed the question Id, context, prompt, multiple-choice choices, correct answer, and the specific subject area within Nephrology to which the question pertained. For all automated models (Koala, Vicuna, Falcon, Orca), we concatenated ``Context," ``Question," and ``Choices (Pick One)" for each forward pass of the model to provide further clarification due to the large input token sizes. If a definitive answer was not provided, we considered the choice incorrect. The latter only occurred with the open-source LLMs.

\subsection*{Experiments}
 We utilized Google Colab's cloud computing GPUs\cite{carneiro2018performance} for some of the LLMs we deployed in this research. Specifically, we used an NVIDIA T4 Tensor Core GPU to execute inference on Koala, while utilizing a more robust NVIDIA A6000 RTX to handle the more memory-heavy Vicuna, Orca, and Falcon 40B model. To perform inference on Koala, Orca, and Vicuna, we utilized the LLaMA tokenizer, which is a statistical NLP technique used to divide a text corpus into smaller, predefined ``tokens." We utilized the LLaMAForCausalLM model from the HuggingFace\cite{wolf2020huggingfaces} library to load the subsequent models.  We configured Koala with the bitsandbytes accelerate package\cite{dettmers20228bit,dettmers2022llmint8}, and loaded it in an 8-bit format to reduce computational costs and enable faster inference times. In order to configure Koala's runtime, we designated the use case as text generation and assigned a temperature of 0.7 to limit the randomness and diversity of the model's predictions. Additionally, we set the top\_p parameter to 0.95, ensuring that Koala only selects word choices with the highest probability during text generation. To prevent Koala from outputting repetitive text, we also applied a repetition penalty parameter of 1.15. To run Vicuna locally, we downloaded the ggml-vic13b-q5\_1.bin file from Hugging Face and ran the model locally using LLaMA\_cpp\_python, a Python wrapper that enables us to utilize the helpful tools from LLaMA.cpp in our Python notebooks. We deployed both the Koala 7B and Vicuna 13B models in a similar manner, utilizing the same LLaMA tokenizer and LLaMA base model. We loaded Samwit's koala-7b and TheBloke’s stable-vicuna-13B-HF models from HuggingFace also with 8-bit precision as extensions of the base LLaMA model. To run Falcon, we utilized the instruction-tuned tiiuae/falcon-7b-instruct\cite{falcon40b} model from HuggingFace. To run this model, we imported the AutoTokenizer and AutoModelForCausalLM functions from the transformers library. The data type of the pipeline was torch.float16. An issue that we observed with Falcon was that it often generated an empty response to the nephSAP questions. To address this issue, we adjusted the model hyperparameters to min\_length=5, max\_length=800, and top\_k=10. We then implemented a while loop based on the condition that the output is not empty. Within the loop, we incremented all hyperparameter values by five until a response was obtained. Lastly, for Orca-mini, we utilized the provided hyperparameters from Hugging Face, with the following settings: top\_p set to 1.0, temperature set to 0.7, top\_k set to 50, and a generated length of 1024 tokens.
\subsection*{Model Metrics and Evaluation}

 After running inference on the models, we obtain a csv file containing predicted answers for each ID. We then parsed the text file of answers (answers.txt) to match each predicted answer with a ground truth response. Our primary objective is to determine the raw accuracies of each model for the multiple-choice answers. To do this, we developed a script to parse the output based on correct input answers. This was necessary due to the large volume of multiple-choice questions, making manual review and comparison impractical. The primary challenge was the variability in outputs generated by different language models, resulting from their training on different datasets and sentence structures. Our approach involved analyzing outputs to identify common patterns across models. By examining numerous samples, we discovered consistent structural elements within the responses. We coded a script to recognize these patterns as regular expressions and extract matching correct answers. We utilized regular expressions to define specific formatting criteria and extract relevant portions of the outputs. This allowed us to identify the responses that aligned with the correct answers, even with slight variations. To ensure accuracy and reliability, we implemented a comprehensive error-checking mechanism. This involved checks for linguistic consistency, contextual relevance, and logical coherence. By validating the extracted answers against these criteria, we filtered out discrepancies or errors, enhancing the overall correctness of the evaluations. The automated comparison checker streamlined the evaluation process, eliminating time-consuming manual reviews. 

\subsection*{Experimental Evaluation}
In all models we evaluated the percentage of the questions that were correctly answered as an informative metric to assess their ability. In addition, in the open-source LLMs we conducted a deeper analysis of the quality and semantic meaning of their responses to the questions to determine which LLM had the highest potential for future fine-tuning. We utilized the BiLingual Evaluation Understudy\cite{papineni2002bleu}(BLEU) metric, from the NLTK toolkit\cite{loper2002nltk}.  BLEU is commonly used in machine translation problems, for example, to determine the language translation quality of a NLP model. The BLEU metric gives a score from 0 to 1, with 0 indicating that the LLM's generated text is of extremely poor quality compared to the ground truth, and a score of 1 indicating that the generated text is most similar to the ground truth. 

Another metric we employed was the word error rate (WER)\cite{ali2018word} from the jiwer package\cite{morris2004and}. WER is employed in machine translation problems and is parameterized as (Substitutions + Insertions + Deletions)/Total Words to assess differences between ground truth explanations and explanations given by the LLMs. 

We furthermore implemented the cosine similarity metric\cite{mikolov2013efficient} from sklearn, which outputs a score between 0 and 1, representing similarity in strings of text based on word embeddings (LLM vs ground truth explanation). While the BLEU, WER, and cosine similarity scores provide a general indication of text similarity, we also included the number of questions for which each LLM achieved a score $\geq$ to 0.5 for both the BLEU and cosine scores. We reported the overall score ± the standard deviation. A BLEU $\geq$ 0.5 is a high-quality translation\cite{BLEUWINDOWS}, and a WER of 5-10\% is considered good quality\cite{WERWINDOWS}.
\section*{Results}
\begin{table}
\centering
\caption{Comparison of the Overall Correct Responses Among the LLMs (858 questions)}
\begin{tabularx}{0.8\textwidth}{>{\centering\arraybackslash}X>{\centering\arraybackslash}X>{\centering\arraybackslash}X}
\textbf{LLM} & \textbf{Number Correct} & \textbf{Percent Correct (\%)} \\ \hline
GPT-4        & 629                     & 73.3                          \\
Claude 2     & 467                     & 54.4                          \\
Vicuna       & 219                     & 25.5                          \\
Orca-Mini    & 147                     & 17.1                          \\
Falcon       & 155                     & 18.1                          \\
Koala        & 204                     & 23.8                          \\ \hline
\end{tabularx}
\end{table}

Table 1 shows the test-taking ability of the different LLMs in answering the Nephrology subspecialty nephSAP multiple-choice questions. Among the open-source models listed, Vicuna achieved the highest score of 25.6\%. Koala had the second-highest score with 23.8\% of the questions correctly answered. Falcon achieved a score of 18.1\% while Orca-mini had the lowest score of 17.1\%. Taking into consideration the number of questions and the choices per question (which varied), we calculated that had the answers been chosen randomly, a score of 23.8\% would have been expected statistically. Among the open-sourced LLMs, only Koala achieved a score slightly above this level. In contrast, GPT-4 did significantly better with a score of 73.3\%, whereas Claude 2, although more successful than the open-sourced LLMs, did more poorly than GPT-4 with a score of 54.4\%. A passing grade on nephSAP questions for human test takers is 75\%. In addition to assessing the overall score on all 858 questions, we broke down the LLM answer choices further based on individual nephSAP question topics within Nephrology and scored each of the topics separately for each LLM. The results are shown in Figure 2.  The open-source LLMs did very poorly in all Nephrology topics. In general, Claude 2 obtained a higher score on all areas in Nephrology while achieving a passing grade in one of the topics.  GPT-4 did exceptionally well and achieved human-like performance in the majority of topics. 

The majority of open-source LLMs achieved an overall score that did not differ from what would be expected had the questions been answered randomly. To further assess the poorly scoring open-sourced models, we further evaluated the quality of their answer explanations. 
As depicted in Figure 3, all of the open-sourced LLMs achieved scores within a reasonable range for Word Error Rate (WER) between 0 to 10\%. However, it is evident that all of the models exhibited sub-optimal performance for the BLEU and cosine similarity score metrics. Specifically, only 9\% of the explanations generated by Vicuna had a BLEU score $\geq$ 0.5, 8\% of the explanations by Falcon met the criteria of a BLEU score $\geq$ 0.5, Orca-mini achieved this threshold for 5\% of its explanations, and lastly, Koala reached only 5\% of explanations with a BLEU score $\geq$ 0.5. In the cosine similarity score, all models exhibited sub-optimal scores above 0.5 (Figure 3). Orca-mini achieved the highest score with 254 questions having a cosine similarity score above 0.5. Vicuna was next best with 243 questions surpassing the 0.5 threshold for cosine similarity. Falcon scored 201, and finally, Koala obtained a score of 181 questions with an embedding greater than or equal to 0.5. Overall, the results further demonstrate the poor test-taking ability of these open-sourced LLMs on nephSAP questions.

\begin{figure}[t]
\centering
\includegraphics[width=1\linewidth]{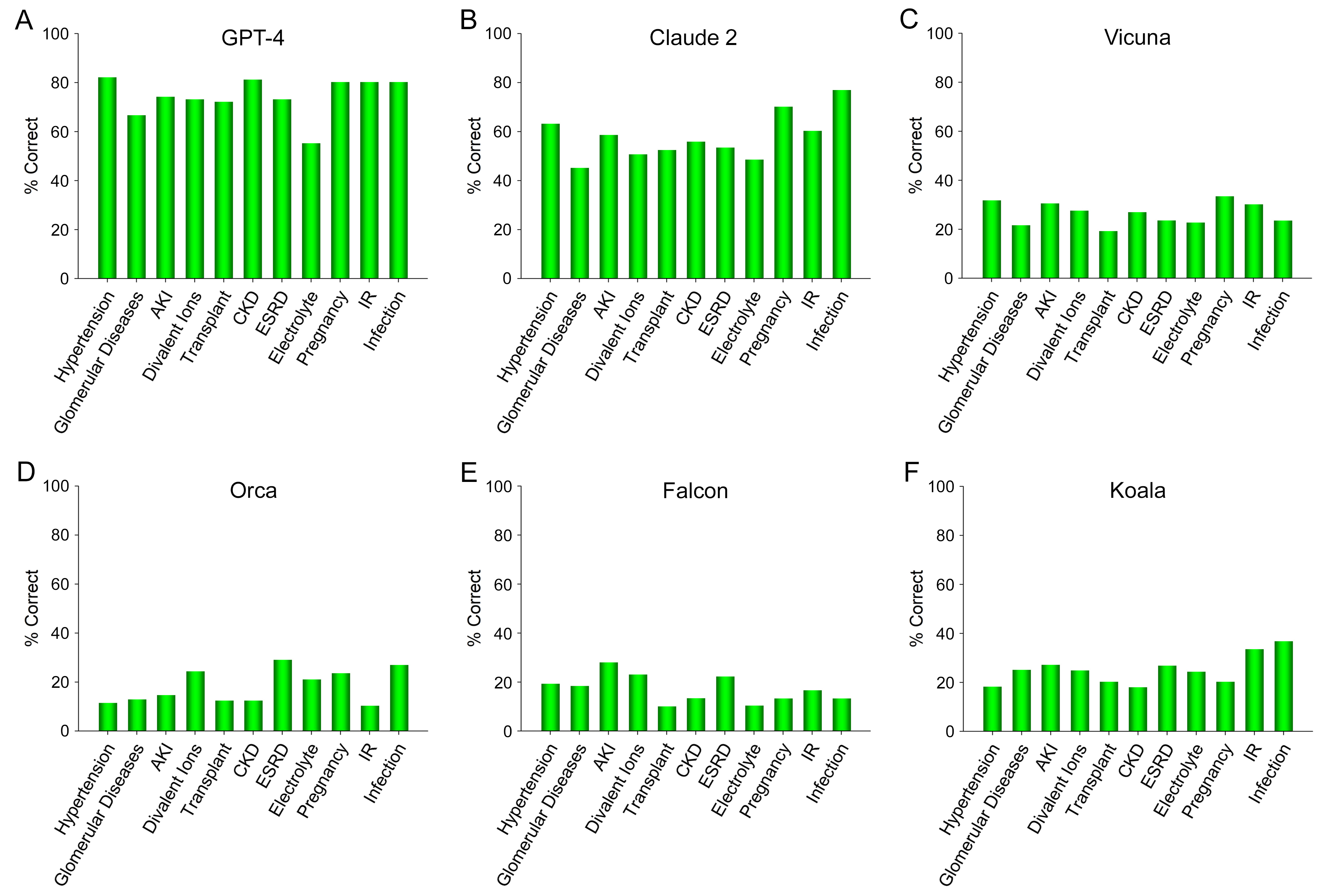}
\caption{Percentage of the correct answers among the various LLMs for each topic within Nephrology.}
\label{fig:stream}
\end{figure}

\begin{figure}[t]
\centering
\includegraphics[width=1\linewidth]{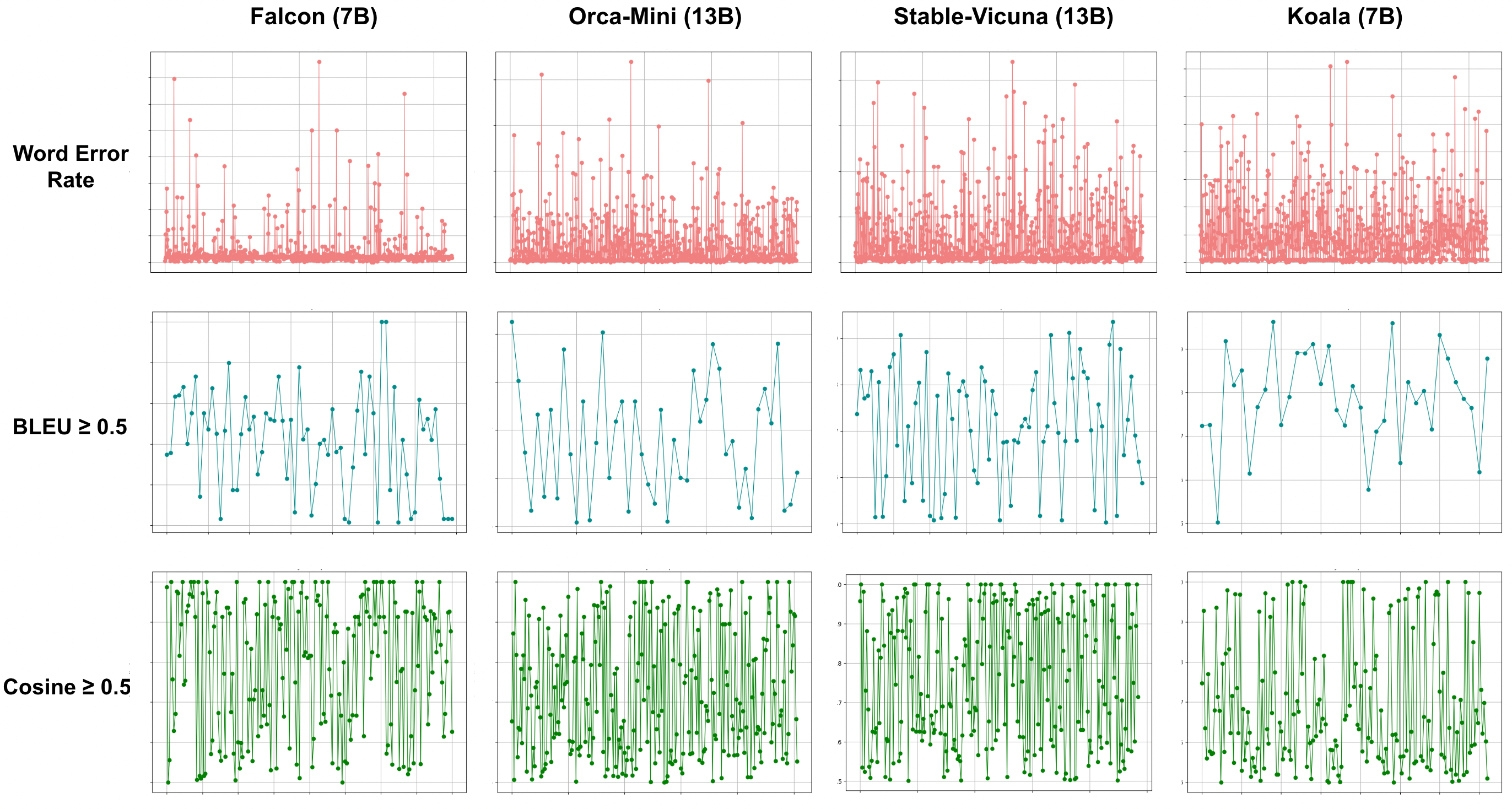}
\caption{Visualization of overall Word Error Rates, BLEU, and Cosine scores. Observe the remarkably low word error rate for Falcon, as well as the fact that Falcon, Orca-Mini, and Stable-Vicuna have numerous questions that achieved a score of $\geq$ 0.5 in both BLEU and Cosine.
}
\label{fig:stream}
\end{figure}

\section*{Discussion}

In this study, we analyzed the ability of open-sourced LLM models, including Koala, Orca-Mini, Falcon, and Stable-Vicuna
to answer correctly multiple-choice nephSAP Nephrology test questions compared to GPT-4 and Claude
2 models. Of all the models GPT-4 was clearly superior with 73.3\% of the questions answered correctly. In addition, GPT-4
scored better in all Nephrology topics assessed individually. Claude 2 achieved the second-best results with an overall score of
54.4\%. In comparison to GPT-4 and Claude 2, the open-source models did poorly in total correct answers and in the quality of their explanations. 

There are several potential reasons for our findings. GPT-4 is reported to have been trained on 1.7 trillion parameters that is approximately 130 times larger than the
open-sourced LLM counterparts. Claude 2 was trained on 860 million parameters. Another distinction which has not received
wide spread attention but that we predict will become an increasingly important issue, is that GPT-4 and Claude 2 was trained not only on
publicly available data, but also on third-party data. The open-source LLMs were trained on publicly available data i.e., ShareGPT, WebGPT, Reddit, PubMed, StackExchange and GitHub. It is likely that
the bulk of high-quality data for training LLMs in the medical field and potentially in most knowledge fields resides in
copyrighted material that has been curated and peer-reviewed and is not freely available publicly e.g., textbooks, published articles and
curated datasets. Without negating the importance of the computational power of specific LLMs,
the lack of free access to training data material that is currently not in public domain will likely remain one of the obstacles
to achieving further improved performance for the foreseeable future.

Another aspect that needs to be considered in attaining better results in all models is the need for domain-specific fine-tuning\cite{lv2023parameter}. Models trained without specific optimization in specialized knowledge domains may yield suboptimal results on domain specific questions\cite{zhao2023survey}. In addition, a more complex model of the world that includes cause-effect and temporal-spatial
understanding is currently lacking in LLMs \cite{hobbhahn2022investigating}. Accordingly, enhancing parameter optimization, utilizing diverse and
representative training datasets, incorporating domain-specific fine-tuning, and improving reasoning capabilities are areas for
future research that could potentially result in even better LLM performance capabilities in the medical field and other knowledge
areas.

In medicine, the amount and complexity of the information human doctors need to master increases as they transition from
medical school to internal medicine residency to ultimately subspecialty practice. Based on this simple premise, it would be predicted that an LLM would do more poorly on an internal medicine subspecialty test than on a test for medical students. Accordingly, the success of GPT-4 on the nephSAP Nephrology questions is in our view truly remarkable, with its overall score of 73.3\% (Table 1). Moreover, GPT-4 achieved a score of $\geq$72\% in 9 of 11 nephSAP Nephrology question topics (Figure 2). The passing grade for each topic for human test takers is 75\% with a total of two attempts for each question. Although we did not formally compare the capabilities of earlier GPT models in our study, with regards to subspecialty internal medicine, a recent study of the performance of ChatGPT and GPT-4 on the 2021 and 2022 American College of Gastroenterology self-assessment multiple-choice questions was considered sub-optimal\cite{suchman2022chat}. Finally, a recent study using a subset of KSAP (Kidney Self-Assessment Program) and nephSAP questions on the topic of glomerular disease showed that ChatGPT underperformed \cite{miao2023assessing}.

In analyzing the ability of GPT-4 in each of the individual Nephrology topics, the electrolyte questions received the lowest
score of 55.2\%. This is not surprising since of all topics in Nephrology, the area of fluid and electrolytes/acid-base diagnosis is
equally more of a challenge for human trainees because of its conceptual-quantitative content where reliance on mere
memorization of facts is not sufficient. There are a few reasons to highlight as to why GPT-4 might also have done less well on
questions related to this topic. The lower quality of publicly available material on this topic may be one issue. In addition, LLMs
generate text based on probabilities for the next token, and while some LLMs use code generation to aid in solving problems
that require quantitative reasoning, improving their inherent quantitative understanding remains an active area of research
\cite{lewkowycz2022solving}. It is expected that domain-specific training using curated datasets involving the more complex topics in
Nephrology will help to improve LLM performance in the future.

Our results in Nephrology show clear evidence that GPT-4 has achieved human-like test-taking ability in one of the difficult
subspecialty fields of internal medicine. The capabilities of AI models and the number of applications in medicine they will be used for
will only increase in the future. We have clearly entered a new era where large healthcare systems, medical schools and their
departments will increasingly embrace these AI models because of perceived or actual cost-saving opportunities and
efficiencies that will be afforded. Moreover, the increasing opportunities for individual personalized internal medicine
subspecialty training that includes AI creation of multiple-choice questions, adaptive learning that takes into consideration
gaps in one's knowledge, digital doctor-patient simulated interactions and AI physician copilots to aid in patient care are only
some of the areas where the field of subspecialty internal medicine including Nephrology will experience profound changes
in the future.

\section*{Acknowledgments}
Ira Kurtz is supported in part by funds from the Smidt Family Foundation, the Factor Family Foundation, the Davita Allen Nissenson Research Fund and the Ralph Block Family Foundation.

\bibliography{sample.bib}

\end{document}